# A Rapid Physics-Informed Machine Learning Framework Based on Extreme Learning Machine for Inverse Stefan Problems

Professor Pei-Zhi **Zhuang**
E-mail: zhuangpeizhi@sdu.edu.cn
School of Qilu Transportation, Shandong University, Jinan, 250002, China

Mrs Ming-Yue **Yang**
E-mail: yangmingyuesdu@mail.sdu.edu.cn
School of Qilu Transportation, Shandong University, Jinan, 250002, China

Dr Fei **Ren**
Email: ren87@outlook.com
School of Mechanical Engineering, Qilu University of Technology (Shandong Academy of Sciences), Jinan, 250353, China

Dr Hong-Ya **Yue**
Email: yuehongya@mail.sdu.edu.cn
School of Qilu Transportation, Shandong University, Jinan, 250002, China

Dr He **Yang**
Email: yanghesdu@mail.sdu.edu.cn
Corresponding author
School of Qilu Transportation, Shandong University, Jinan, 250002, China

## Abstract

The inverse Stefan problem, as a typical phase-change problem with moving boundaries, finds extensive applications in science and engineering. Recent years have seen the applications of physics-informed neural networks (PINNs) to solving Stefan problems, yet they still exhibit shortcomings in hyperparameter dependency, training efficiency, and prediction accuracy. To address this, this paper develops a physics-informed extreme learning machine (PIELM), a rapid physics-informed learning method framework for inverse Stefan problems. PIELM replaces conventional deep neural networks with an extreme learning machine network. The input weights are fixed in the PIELM framework, and the output weights are determined by optimizing a loss vector of physical laws composed by initial and boundary conditions and governing partial differential equations (PDEs). Then, solving inverse Stefan problems is transformed into finding the Moore–Penrose generalized inverse by the least squares method. Case studies show that the PIELM can increase the prediction accuracy by 3~7 order of magnitude in terms of the relative $L_2$ error, and meanwhile saving more than 94% training time, compared to conventional PINNs.

## 1. Introduction

Stefan problems origin from the phase change of ice melting to water with a heat source, where a moving boundary exists at the ice-water (i.e., solid-liquid) interface (Sproull 1962; Hanzawa 1981). Investigation of Stefan problems has already been a popular branch in mathematics and computational methods, and the model of Stefan problems has also been applied to various fields such as materials science engineering (Friedman and Hu 1996; Zhang et al. 2021), fluid dynamics and interface tracking (Kim et al. 2019; Cukrov et al. 2021), heat conduction and phase change (Choi and Kim 2012), biomedical engineering (Choi and Kim 2012), among others (Tarzia 2000).

Due to the existence of dynamic phase-change boundary, solving Stefan problems is not as easy as solving classical PDEs (Voller et al. 2004; Briozzo et al. 2007; Liu et al. 2012; Zhou and Xia 2015). Several analytical and numerical methods are available in the literature (Voller 1987; Womble 1989; Date 1992; Unverdi and Tryggvason 1992), and their limitations for various Stefan problems were discussed by Gupta (2017), Wang and Perdikaris (2021) and Ren et al. (2025). Recently, physics-informed machine learning has become increasingly popular due to the rapid development of artificial intelligence and computational science. The advantage of this approach is that it combines the strengths of rigorous physics and available data (Kilkenny and Robinson 2018; Raissi et al. 2019; Yeo and Melnyk 2019; Nabian and Meidani 2020; Guo et al. 2022; Zhang et al. 2022; Sajadi et al. 2025). Incorporating physical knowledge into the model makes it easier to understand the internal mechanisms of machine learning models and meanwhile provide reasonable explanations for the prediction results. One hot-topic branch is the physics-informed neural networks (PINNs) that can solve PDEs with given initial conditions, boundary conditions, and measured data (if any) with the help of neural networks (Berg and Nyström 2018; Raissi et al. 2019; Yang et al. 2020; Cuomo et al. 2022; Sukumar and Srivastava 2022; Song et al. 2024). A few attempts have also been made to solve Stefan problems (Wang and Perdikaris 2021; Li et al.

2023; Kathane and Karagadde 2024; Madir et al. 2025; Park et al. 2025). For example, Wang and Perdikaris (2021) used a dual-network PINN to solve the Stefan problem. Symbat (2025) extended PINNs to a two-dimensional two-phase Stefan model, embedding the solid-liquid phase heat equation and Stefan conditions into the dissipation term, achieving a relative error of approximately $10^{-3}$ in temperature and interface predictions. However, there are a few limitations when using PINNs:

(1) Determination of network hyperparameters is heavily reliant on experience.

(2) The training efficiency is low and several minutes may be required for one-dimensional problems and more for high-dimensional problems.

(3) The accuracy may be in the order of $10^{-3}$ in terms of the relative $L^2$ error, which is not comparable to conventional numerical methods.

To overcome these limitations of PINNs, Dwivedi and Srinivasan (2020) proposed a rapid physics-informed machine learning framework by replacing deep neural networks in PINN with a single-layer extreme learning machine (ELM), which is known as physics-informed extreme learning machine (PIELM). Most recently, Ren et al. (2025) solved Stefan problems using an iterative dual-network PIELM approach. They demonstrated that PIELM can significantly improve the training time and solution accuracy by more than 50 and 1000 times compared to conventional PINNs. However, only the forward Stefan problems are tackled in Ren et al. (2025). In fact, the inverse Stefan problems can be equally important for their wide applications in science and engineering (Dihn 1991; AlKhalidy 1996; Garshasbi and Dastour 2016; Nauryz and Jabbarkhanov 2025; Srisuma et al. 2025). For example, the inverse Stefan problems can be used to determine thermal physical parameters such as thermal conductivity and specific heat capacity of materials when optimizing parameters in metal casting and heat treatment processes (Chen et al. 2015). In steam-assisted gravity drainage technology, the model of inverse Stefan problems can be used to calculate the lateral expansion of the steam cavity to optimize the extraction efficiency of oil wells (Chen et al. 2015). It can also be used to simulate temperature distribution and phase interface changes during cell culture processes to optimize tissue repair techniques (Friedman and Hu 1996). Therefore, solving inverse Stefan problems is necessary but it remains to be explored whether PIELM is still powerful for inverse Stefan problems to facilitate the development of mathematics and computational science.

This paper solves inverse Stefan problems using PIELM method as a companion paper of Ren et al. (2025). Compared to forward Stefan problems, the movement of the phase-change interface is prior-known in the considered inverse problems. This will lead to linear governing equations for inverse Stefan problems, and the ELM network can be trained with more convenience and also faster. The rest of the paper is organized as follows. The preliminaries for inverse Stefan problems and PINN framework is shown in Section 2. The PIELM framework is shown in Section 3. Then the performance of PIELM for inverse Stefan problems is tested in Section 4. Finally, key conclusions are drawn in Section 5.

## 2. Preliminaries Inverse Stefan Problems and PINN framework

### *2.1. Mathematical definition of Inverse Stefan Problems*

The mathematical formulations for Stefan problems are shown by taking ice melting as an example. As shown in Figure 1, $\Omega_1(t)$ and $\Omega_2(t)$ denotes the solid region for the ice and the liquid region for water, respectively. The solid and liquid regions have fixed boundaries $\mathit{\Gamma}_1(\boldsymbol{x})$ and $\mathit{\Gamma}_2(\boldsymbol{x})$, respectively, denote the moving interface between the two regions. The governing PDEs for Stefan problems can be expressed as

$$\frac{\partial u_1(\boldsymbol{x},t)}{\partial t} - k_1\nabla^2 u_1(\boldsymbol{x},t) = 0, \boldsymbol{x} \in \Omega_1(t) \tag{1}$$

$$\frac{\partial u_2(\boldsymbol{x},t)}{\partial t} - k_2\nabla^2 u_2(\boldsymbol{x},t) = 0, \boldsymbol{x} \in \Omega_2(t) \tag{2}$$

where $u_1$ and $u_2$ denotes the temperature in the solid and liquid regions, respectively; $t$ ($\geq 0$) is the time; and $\boldsymbol{x}$ is the spatial coordinate point; for ice melting, $k_1$ and $k_2$ are the thermal diffusivity of ice and water (known parameters); $\nabla^2$ means the Laplace operator.

The initial conditions and the fixed boundary conditions are generally expressed as

$$u_1(\boldsymbol{x},0)=I_1(\boldsymbol{x}),\boldsymbol{x}\in\Omega_1(t) \tag{3}$$

$$u_2(\boldsymbol{x},0)=I_2(\boldsymbol{x}),\boldsymbol{x}\in\Omega_2(t) \tag{4}$$

$$u_1(\boldsymbol{x},t)=B_1(\boldsymbol{x},t),\boldsymbol{x}\in\mathit{\Gamma}_1(\boldsymbol{x}) \tag{5}$$

$$u_2(\boldsymbol{x},t)=B_2(\boldsymbol{x},t),\boldsymbol{x}\in\mathit{\Gamma}_2(\boldsymbol{x}) \tag{6}$$

where $I_1$ and $I_2$ are the initial conditions for solid and liquid, respectively; $B_1$ and $B_2$ are the boundary conditions for solid and liquid, respectively.

To solve inverse Stefan problems, additional initial conditions and boundary conditions are required at the moving boundary

$$\Sigma(0)=\Sigma_0 \tag{7}$$

$$u_1(\boldsymbol{x},t)=u_2(\boldsymbol{x},t)=u_0,\boldsymbol{x}\in\Sigma(t) \tag{8}$$

$$\frac{\partial u_1(\boldsymbol{x},t)}{\partial \boldsymbol{n}}+\xi\frac{\partial u_2(\boldsymbol{x},t)}{\partial \boldsymbol{n}}=g(\boldsymbol{x},t) \tag{9}$$

where $\boldsymbol{n}$ is the unit normal to $\Sigma(t)$ oriented from $\Omega_1(t)$ to $\Omega_2(t)$; $\Sigma_0$, $u_0$, $\xi$ are known constants and $g(\Sigma,t)$ is the known function. For inverse Stefan problems considered in this paper, the moving interface $\Sigma(t)$ is given while the temperature distributions for ice and water, and the boundary conditions at the fixed boundaries should be inferred. We only consider this type of inverse Stefan problems because in this case Eqs. (1) and (2) will become linear PDEs. For other type inverse Stefan problems with unknown moving boundary, iterative PIELM approach is required and can be seen in Ren et al.

(2025).

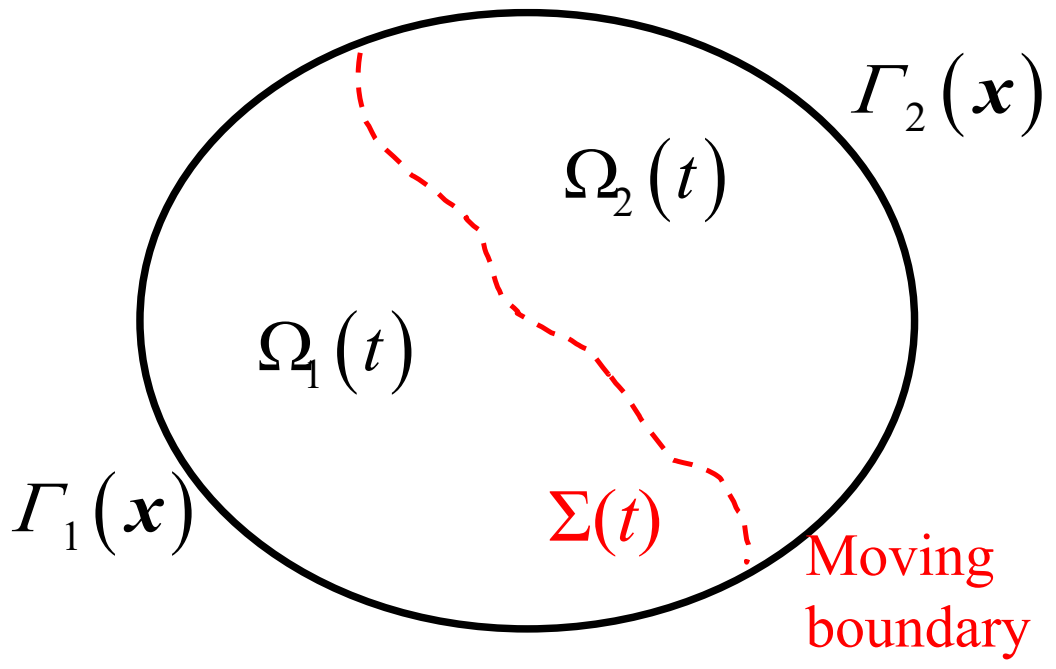

**Figure 1  Schematic of Stefan problems**

### *2.2. PINN Framework for Inverse Stefan Problems*

To solve Stefan problems, Wang and Perdikaris (2021) proposed a dual-net PINN framework to approximate the target solution. Their PINN framework can be simplified as a single-network framework for the present inverse analysis, as the moving boundary is known in advance. As shown in Figure 2, a number of collection points $\{\boldsymbol{x},t\}$ are input into the deep neural network with several hidden layers, where a few neurons in each hidden layer with the activation function $\sigma$. Using initial network parameters (randomly generated), the outputs of the neural network $u_1(\boldsymbol{x},t)$ and $u_2(\boldsymbol{x},t)$ are the latent temperature distribution in the solid and liquid regions, respectively. Obviously, $u_1(\boldsymbol{x},t)$ and $u_2(\boldsymbol{x},t)$ are not accurate enough before training the neural network. Then, the derivatives of $u_1(\boldsymbol{x},t)$ and $u_2(\boldsymbol{x},t)$ in terms of spatial coordinates and time are calculated by automatic differentiation (AD). To derive reasonably accurate predictions, the physical laws of governing equations, initial conditions, and boundary conditions as defined by Eqs. (1)~(9), and are substituted into the total loss function $\mathcal{L}(\boldsymbol{\theta})$ defined as

$$
\begin{aligned}
\mathcal{L}(\boldsymbol{\theta}) &= \lambda_{r,1}\mathcal{L}_{r,1}(\boldsymbol{\theta}) + \lambda_{ic,1}\mathcal{L}_{ic,1}(\boldsymbol{\theta}) + \lambda_{sbc,1}\mathcal{L}_{sbc,1}(\boldsymbol{\theta}) \\
&\quad + \lambda_{r,2}\mathcal{L}_{r,2}(\boldsymbol{\theta}) + \lambda_{ic,2}\mathcal{L}_{ic,2}(\boldsymbol{\theta}) + \lambda_{sbc,2}\mathcal{L}_{sbc,2}(\boldsymbol{\theta}) + \lambda_{sc}\mathcal{L}_{sc}(\boldsymbol{\theta})
\end{aligned}
\tag{10}
$$

$$
\mathcal{L}_{r,1}(\boldsymbol{\theta}) = \frac{1}{N_C}\sum_{i=1}^{N_C}\left|\frac{\partial u_1(\boldsymbol{x}_i, t_i; \boldsymbol{\theta})}{\partial t} - k_1\nabla^2 u_1(\boldsymbol{x}_i, t_i; \boldsymbol{\theta})\right|^2 \tag{11}
$$

$$
\mathcal{L}_{r,2}(\boldsymbol{\theta}) = \frac{1}{N_C}\sum_{i=1}^{N_C}\left|\frac{\partial u_2(\boldsymbol{x}_i, t_i; \boldsymbol{\theta})}{\partial t} - k_2\nabla^2 u_2(\boldsymbol{x}_i, t_i; \boldsymbol{\theta})\right|^2 \tag{12}
$$

$$
\mathcal{L}_{ic,1}(\boldsymbol{\theta}) = \frac{1}{N_I}\sum_{l=1}^{N_I}\left|u_1(\boldsymbol{x}_l, 0; \boldsymbol{\theta}) - I_1(\boldsymbol{x}_l)\right|^2 \tag{13}
$$

$$
\mathcal{L}_{ic,2}(\boldsymbol{\theta}) = \frac{1}{N_I}\sum_{l=1}^{N_I}\left|u_2(\boldsymbol{x}_l, 0; \boldsymbol{\theta}) - I_2(\boldsymbol{x}_l)\right|^2 \tag{14}
$$

$$
\mathcal{L}_{sbc,1}(\boldsymbol{\theta}) = \frac{1}{N_C}\sum_{i=1}^{N_C}\left|u_1(s(t_i), t_i; \boldsymbol{\theta}) - u_s\right|^2 \tag{15}
$$

$$
\mathcal{L}_{sbc,2}(\boldsymbol{\theta}) = \frac{1}{N_C}\sum_{i=1}^{N_C}\left|u_2(s(t_i), t_i; \boldsymbol{\theta}) - u_s\right|^2 \tag{16}
$$

$$
\mathcal{L}_{sc}(\boldsymbol{\theta}) = \frac{1}{N_C}\sum_{i=1}^{N_C}\left|\frac{\partial u_1}{\partial \boldsymbol{n}}(s(t_i), t_i; \boldsymbol{\theta}) + \xi\frac{\partial u_2}{\partial \boldsymbol{n}}(s(t_i), t_i; \boldsymbol{\theta}) - g(s(t_i), t_i)\right|^2 \tag{17}
$$

where $\boldsymbol{\theta}$ is the network parameters for network training; $\{\lambda_{r,1}, \lambda_{r,2}, \lambda_{ic,1}, \lambda_{ic,2}, \lambda_{sbc,1}, \lambda_{sbc,2}, \lambda_{sc}\}$ are the loss weights, and in this paper the self-adaptive method of Xiang et al. (2022) is used to determine these loss weights; $\{x_i, t_i\}_{i=1}^{N_C}$ and $\{x_l, 0\}_{l=1}^{N_I}$ are the training points for governing PDEs and initial conditions; $N_C$ and $N_I$ are the corresponding training point numbers. Note that $\{x_i, t_i\}_{i=1}^{N_C}$ are collected in the whole domain during the training process, and $u_1(\boldsymbol{x}, t)$ and $u_2(\boldsymbol{x}, t)$ are output from solution domain $\Omega_1(t)$ and $\Omega_2(t)$, respectively. $N_C$ points are also adopted for the moving boundary in order to conveniently deal with the compatibility of $u_1(\boldsymbol{x}, t)$ and $u_2(\boldsymbol{x}, t)$ at the moving interface.

Normally, the gradient-descent method is adopted to minimize the total loss, making $\mathcal{L}(\boldsymbol{\theta})$ approach zero and updating $\boldsymbol{\theta}$ in each iteration. After training, the trained neural network can give reasonable predictions for temperature distributions in the solid and liquid regions. However, it is necessary to mention that, the gradient-descent method for training deep neural networks is time-consuming, making conventional PINN less efficient (Dwivedi and Srinivasan 2020; Wang et al. 2022).

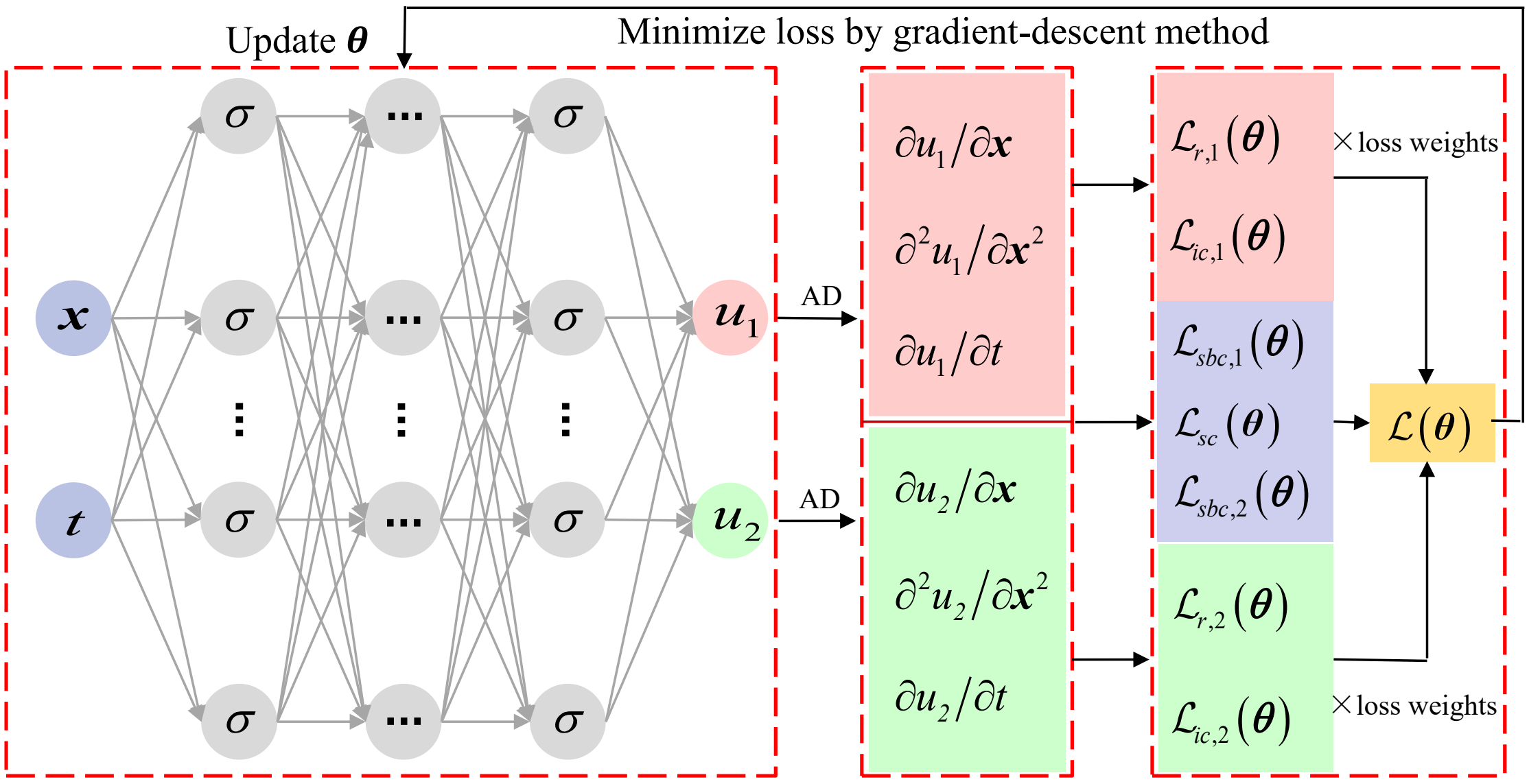

**Figure 2 PINN framework for inverse Stefan problems**

## 3. PIELM framework for Inverse Stefan Problems

### *3.1. Brief Review of ELM*

ELM is a kind of single-hidden-layer feedforward neural network proposed by Huang et al. (2006). An ELM network with $d$ input nodes and $D$ output nodes is illustrated in Figure 3. The input layer consists of $d$ nodes, forming the input vector $\boldsymbol{x}=[x_1,x_2,...,x_d]$. The single hidden layer contains $M$ neurons, each employing an activation function $\sigma$. The input weight matrix $\boldsymbol{w}=[\boldsymbol{w}_1,\boldsymbol{w}_2,...,\boldsymbol{w}_M]\in\mathbb{R}^{d\times M}$ and hidden layer bias vector $\boldsymbol{b}=[b_1,b_2,...,b_M]^{\mathrm{T}}$ in the ELM network are assigned randomly, and are then fixed during the training process. The output of each neuron $h_m$ in hidden layer is calculated by

$$h_m(\boldsymbol{x}) = \sigma\left(\boldsymbol{w}_m^{\mathrm{T}} \boldsymbol{x}^{\mathrm{T}} + b_m\right) \tag{18}$$

where $\boldsymbol{w}_m = \left[w_{m,1}, w_{m,2}, ..., w_{m,d}\right]^{\mathrm{T}}$ for $m$ = 1, 2, …, $M$. The hyperbolic tangent function (tanh) is selected as the activation function $\sigma$ in this paper. The output layer consists of $D$ nodes, forming the output vector $\boldsymbol{y} = [y_1, y_2, ..., y_D]$. We use $\boldsymbol{\theta} = [\boldsymbol{\theta}_1, \boldsymbol{\theta}_2, ..., \boldsymbol{\theta}_D] \in \mathbb{R}^{M \times D}$ to represent the output weight matrix, The final output $y_j$ for each output node is calculated by

$$y_j = \boldsymbol{H}\boldsymbol{\theta}_j \tag{19}$$

where $\boldsymbol{H}(\boldsymbol{x}) = \left[h_1(\boldsymbol{x}), h_2(\boldsymbol{x}), \ldots, h_M(\boldsymbol{x})\right]$ and $\boldsymbol{\theta}_j = \left[\theta_{j,1}, \theta_{j,2}, ..., \theta_{j,M}\right]^{\mathrm{T}}$ for $j$ = 1, 2, …, $D$. Finally, Eq. (20) encapsulates the operation of ELM, demonstrating how inputs are transformed through the ELM network to generate the outputs.

$$\boldsymbol{y} = \boldsymbol{H}\boldsymbol{\theta} \tag{20}$$

The optimal output weight matrix $\boldsymbol{\theta}^*$ can be directly obtained by using the Moore-Penrose generalized inverse matrix of $\boldsymbol{H}$, as $\boldsymbol{\theta}^* = (\boldsymbol{H}^{\mathrm{T}}\boldsymbol{H})^{-1}\boldsymbol{H}^{\mathrm{T}}\boldsymbol{y}$. Once $\boldsymbol{\theta}^*$ is determined, the target outputs, $\boldsymbol{y}$, will be derived by $\boldsymbol{y} = \boldsymbol{H}\boldsymbol{\theta}^*$.

It can be found that the backpropagation of errors is not required in the ELM network. Therefore, compared with traditional neural networks, ELM offers advantages including fast training speed, non-iterative learning, and minimal parameter tuning.

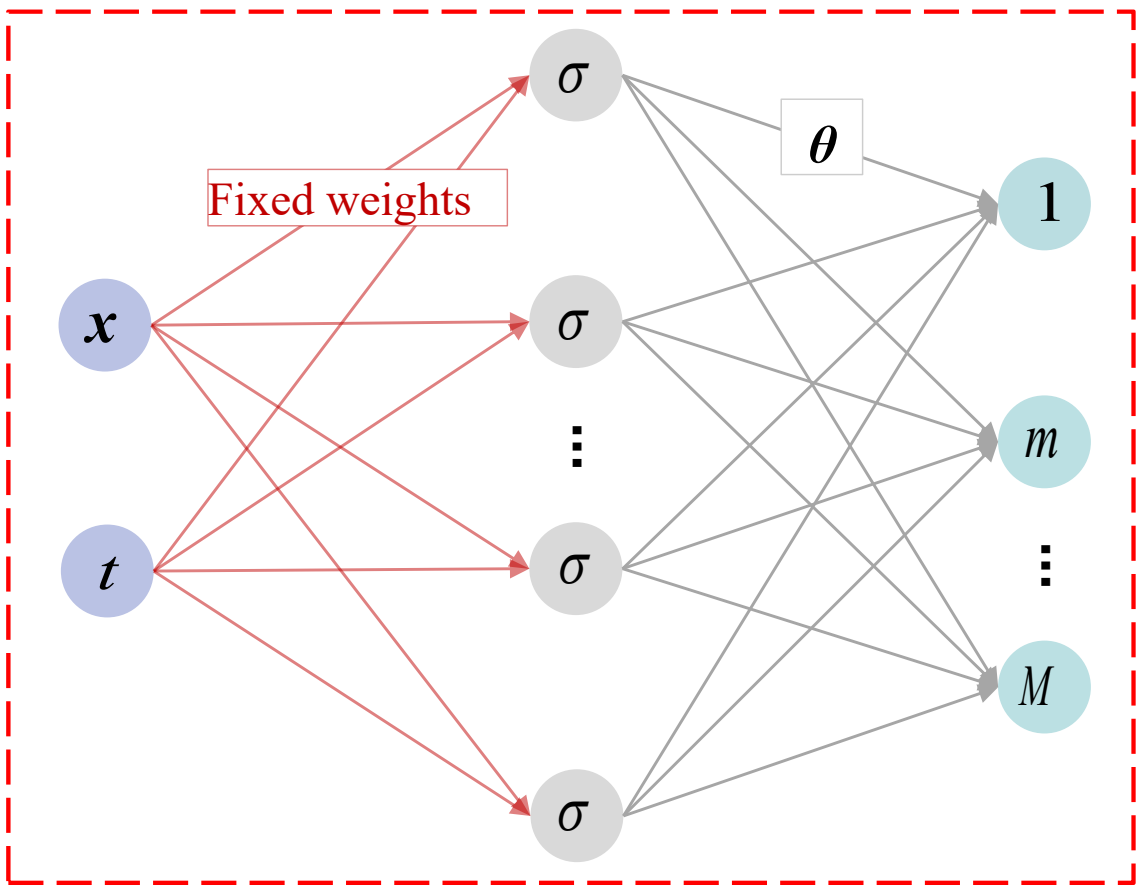

**Figure 3 ELM network architecture**

### *3.3. PIELM Framework*

PIELM is a fast version of PINN Dwivedi and Srinivasan (2020); (Calabrò et al. 2021; Dong and Li 2021; Schiassi et al. 2021; Liu et al. 2023; De Florio et al. 2024), where a deep neural network is replaced by an ELM. The PIELM architecture for the inverse Stefan problems is shown in Figure 4. The inputs $\{x_i,t_i\}_{i=1}^{N_C}$ are fed into the hidden layer with several neurons, and the temperature distributions $\{u_1, u_2\}$ in the solid and liquid regions are the target outputs. The output weights are calculated by the least squares method by optimizing the loss vector

$$\mathcal{L}_C^{(i)}(\boldsymbol{\theta})=\left[\mathcal{L}_{r,1}^{(i)}(\boldsymbol{\theta}),\mathcal{L}_{r,2}^{(i)}(\boldsymbol{\theta})\right]^{\mathrm{T}},i=1,2,\ldots,N_C \tag{21}$$

$$\mathcal{L}_I^{(l)}(\boldsymbol{\theta})=\left[\mathcal{L}_{uic,1}^{(l)}(\boldsymbol{\theta}),\mathcal{L}_{uic,2}^{(l)}(\boldsymbol{\theta})\right]^{\mathrm{T}},l=1,2,\ldots,N_I \tag{22}$$

$$\mathcal{L}_B^{(i)}(\boldsymbol{\theta})=\left[\mathcal{L}_{sbc,1}^{(i)}(\boldsymbol{\theta}),\mathcal{L}_{sbc,2}^{(i)}(\boldsymbol{\theta}),\mathcal{L}_{sc}^{(i)}(\theta)\right]^{\mathrm{T}},i=1,2,\ldots,N_C \tag{23}$$

where $\mathcal{L}_C$, $\mathcal{L}_I$ and $\mathcal{L}_B$ represent the loss terms regarding the points for PDEs, initial conditions and moving boundary conditions, respectively, expressed as

$$\mathcal{L}_{r,1}^{(i)}(\boldsymbol{\theta})=\frac{\partial u_1}{\partial t}(x_i,t_i;\boldsymbol{\theta})-k_1\nabla^2 u_1(x_i,t_i;\boldsymbol{\theta}) \tag{24}$$

$$\mathcal{L}_{r,2}^{(i)}(\boldsymbol{\theta})=\frac{\partial u_2}{\partial t}(x_i,t_i;\boldsymbol{\theta})-k_2\nabla^2 u_2(x_i,t_i;\boldsymbol{\theta}) \tag{25}$$

$$\mathcal{L}_{uic,1}^{(l)}(\boldsymbol{\theta}) = u_1(x_l, 0; \boldsymbol{\theta}) - h_1(x_l) \tag{26}$$

$$\mathcal{L}_{uic,2}^{(l)}(\boldsymbol{\theta}) = u_2(x_l, 0; \boldsymbol{\theta}) - h_2(x_l) \tag{27}$$

$$\mathcal{L}_{sbc,1}^{(i)}(\boldsymbol{\theta}) = u_1(s(t_i), t_i; \boldsymbol{\theta}) - u_s \tag{28}$$

$$\mathcal{L}_{sbc,2}^{(i)}(\boldsymbol{\theta}) = u_2(s(t_i), t_i; \boldsymbol{\theta}) - u_s \tag{29}$$

$$\mathcal{L}_{sc}^{(i)}(\boldsymbol{\theta}) = \frac{\partial u_1}{\partial n}(s(t_i), t_i; \boldsymbol{\theta}) + \xi \frac{\partial u_2}{\partial n}(s(t_i), t_i; \boldsymbol{\theta}) - g(s(t_i), t_j) \tag{30}$$

The total loss function $\mathcal{L}(\theta)$ is a vector-valued function representing the physical laws of PDEs and initial and boundary conditions, given by:

$$\mathcal{L}(\boldsymbol{\theta}) = \left[\mathcal{L}_C^{(1)}(\boldsymbol{\theta}), \ldots, \mathcal{L}_C^{(N_C)}(\boldsymbol{\theta}), \mathcal{L}_I^{(1)}(\boldsymbol{\theta}), \ldots, \mathcal{L}_I^{(N_I)}(\boldsymbol{\theta}), \ \mathcal{L}_B^{(1)}(\boldsymbol{\theta}), \ldots, \mathcal{L}_B^{(N_C)}(\boldsymbol{\theta})\right]^{\mathrm{T}} \tag{31}$$

Now the inverse Stefan problem has been converted into solving a set of linear equations by optimizing $\mathcal{L}(\boldsymbol{\theta})=0$, and $\boldsymbol{\theta}$ can be directly solved using the least squares method following the procedures in Section 3.1.

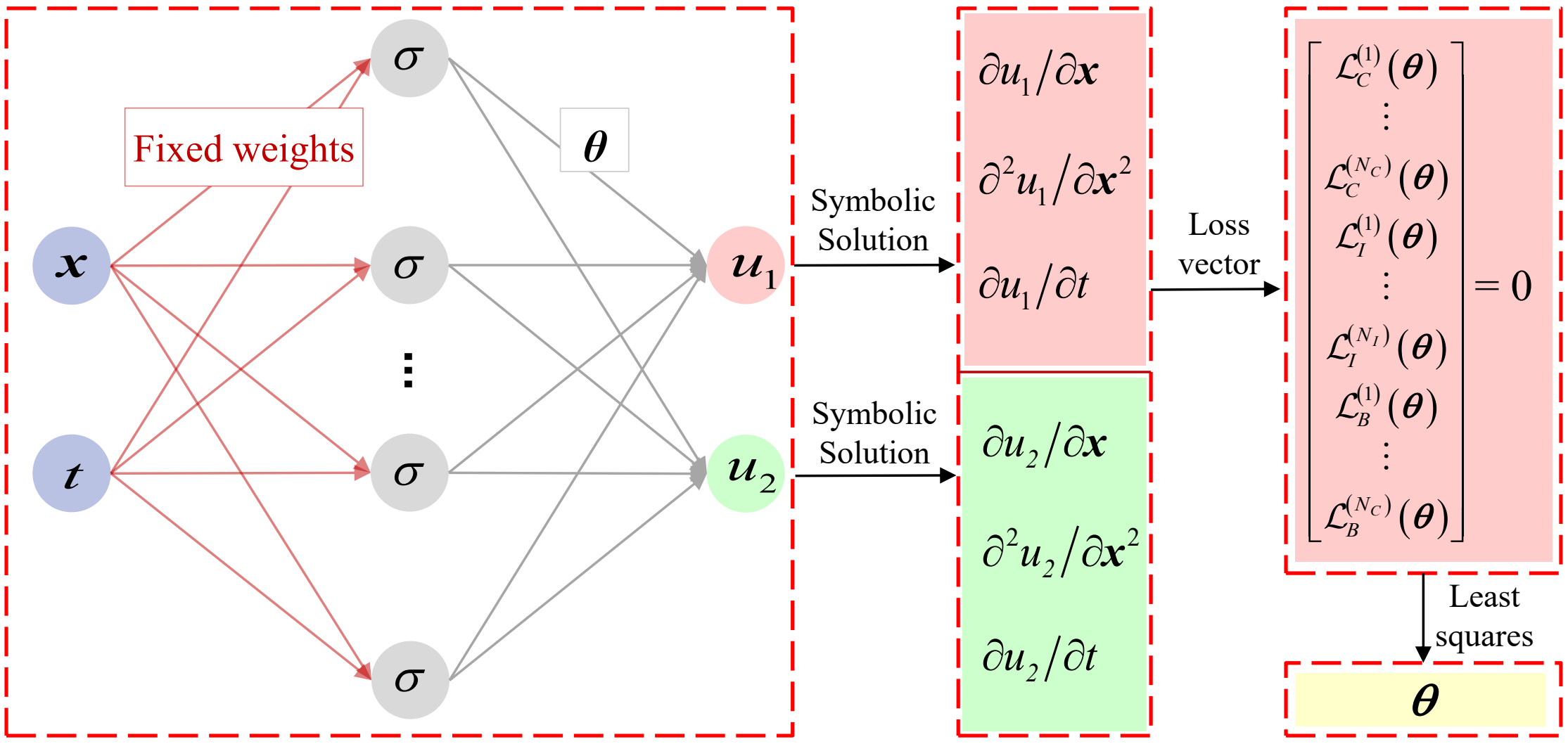

**Figure 4  PIELM framework for inverse Stefan problems**

## 4. Case studies

In this section, four case studies are conducted to show the performance of PIELM for inverse Stefan problems, including:

(i) Case 1: One-dimensional one-phase inverse Stefan problem;

(ii) Case 2: One-dimensional two-phase inverse Stefan problem;

(iii) Case 3: Two-dimensional one-phase inverse Stefan problem;

(iv) Case 4: Two-dimensional two-phase inverse Stefan problem.

The predicted results by the PINN framework are also provided for the purpose of comparison. In all four cases, the hidden layer structures are the same, and the hyperparameters are summarised in Table 1. All the results are calculated by MATLAB software on a computer with an i7-9700 CPU and a 16GB RAM. To better compare the accuracy, the relative $L_2$ error is used as the index, defined as

$$L_2 = \frac{\sqrt{\sum_{1}^{N_{\text{test}}} (\text{exact} - \text{predicted})^2}}{\sqrt{\sum_{1}^{N} (\text{exact})^2}} \tag{32}$$

where $N_{\text{test}}$ is the number of test points. The relative $L_2$ errors and training times for the four cases are summarised in Table 2.

**Table 1 Hyperparameters in PINN and PIELM frameworks**

| Hyperparameters | One-dimensional | | Two-dimensional | |
|---|---|---|---|---|
| | PINN | PIELM | PINN | PIELM |
| Hidden layer Number | 3 | 1 | 3 | 1 |
| Hidden neurons Number | 100 | 150 | 200 | 200 |
| Activation function | tanh | tanh | tanh | tanh |
| Initialization | He | Randomly generated in [-1,1] | He | Randomly generated in [-1,1] |
| Optimizer | ADAM | Least squares | ADAM | Least squares |
| $N_C$ | 1024 | 100 | 10000 | 2000 |
| $N_I$ | 101 | 50 | 4001 | 401 |
| Batch size | 128 | - | 128 | - |
| Learning rate | 0.001 | - | 0.001 | - |
| Epoch | 6000 | - | 5000 | - |

**Table 2 Summary of relative L2 errors and training times**

| Case studies | PIML approaches | $L_2$ | Training time (s) |
| --- | --- | --- | --- |
| Case 1 | PINN | 7.70e-4 | 316 |
| | PIELM | **2.07e-11** | **1.27** |
| Case 2 | PINN | 6.05e-3 | 403 |
| | PIELM | **4.88e-6** | **1.83** |
| Case 3 | PINN | 1.99e-1 | 204 |
| | PIELM | **5.97e-5** | **12.15** |
| Case 4 | PINN | 4.28e-2 | 411 |
| | PIELM | **2.71e-5** | **21.60** |

### *4.1. Case 1: One-dimensional One-phase Inverse Stefan Problems*

Following Wang and Perdikaris (2021), a one-dimensional one-phase inverse Stefan problem in Furzeland (1980) is used as a benchmark for PIELM. The PDE, initial conditions and boundary conditions are

$$\frac{\partial u}{\partial t} - \frac{\partial^2 u}{\partial x^2} = 0, \quad 0 \le x \le s(t) \quad \text{and} \quad 0 \le t \le 1 \tag{33}$$

$$u(x,0) = -\frac{x^2}{2} + 2x - \frac{1}{2}, \quad 0 \le x \le s(0) \tag{34}$$

$$\frac{\partial u}{\partial x}(0,t) = 2, \quad 0 \le t \le 1 \tag{35}$$

It is assumed that the prior-known moving boundary satisfies

$$s(t) = 2 - \sqrt{3-2t} \tag{36}$$

The initial and boundary conditions at the moving boundary are expressed as

$$u(s,t) = 0, \quad 0 \le t \le 1 \tag{37}$$

$$\frac{\partial u}{\partial x}(s,t) = \sqrt{3-2t}, \quad 0 \le t \le 1 \tag{38}$$

The solution for Eqs. (33)~(38) can refer to Wang and Perdikaris (2021) as

$$u(x,t) = -\frac{x^2}{2} + 2x - \frac{1}{2} - t, \quad 0 \le x \le s(t) \tag{39}$$

$$u(0,t) = -\frac{1}{2} - t, 0 \le t \le 1 \tag{40}$$

$$\frac{\partial u}{\partial x}(0,t) = 2, 0 \le t \le 1 \tag{41}$$

Figure 5 shows the temperature distribution and boundary conditions predicted by PINN and PIELM with absolute point-wise error. It shows that both PINN and PIELM provide rather accurate solutions for inverse Stefan problems. In terms of the point-wise error, however, the accuracy of PIELM can reach the order of $10^{-11}$, i.e., $10^{7}$ higher than that of PINN, reaching the order of $10^{-11}$. The training time required by PIELM is also significantly reduced, approximately from 300s to 1s. The comparison of $L_2$ and training time proves the much higher accuracy and efficiency of PIELM for the inverse Stefan problem.

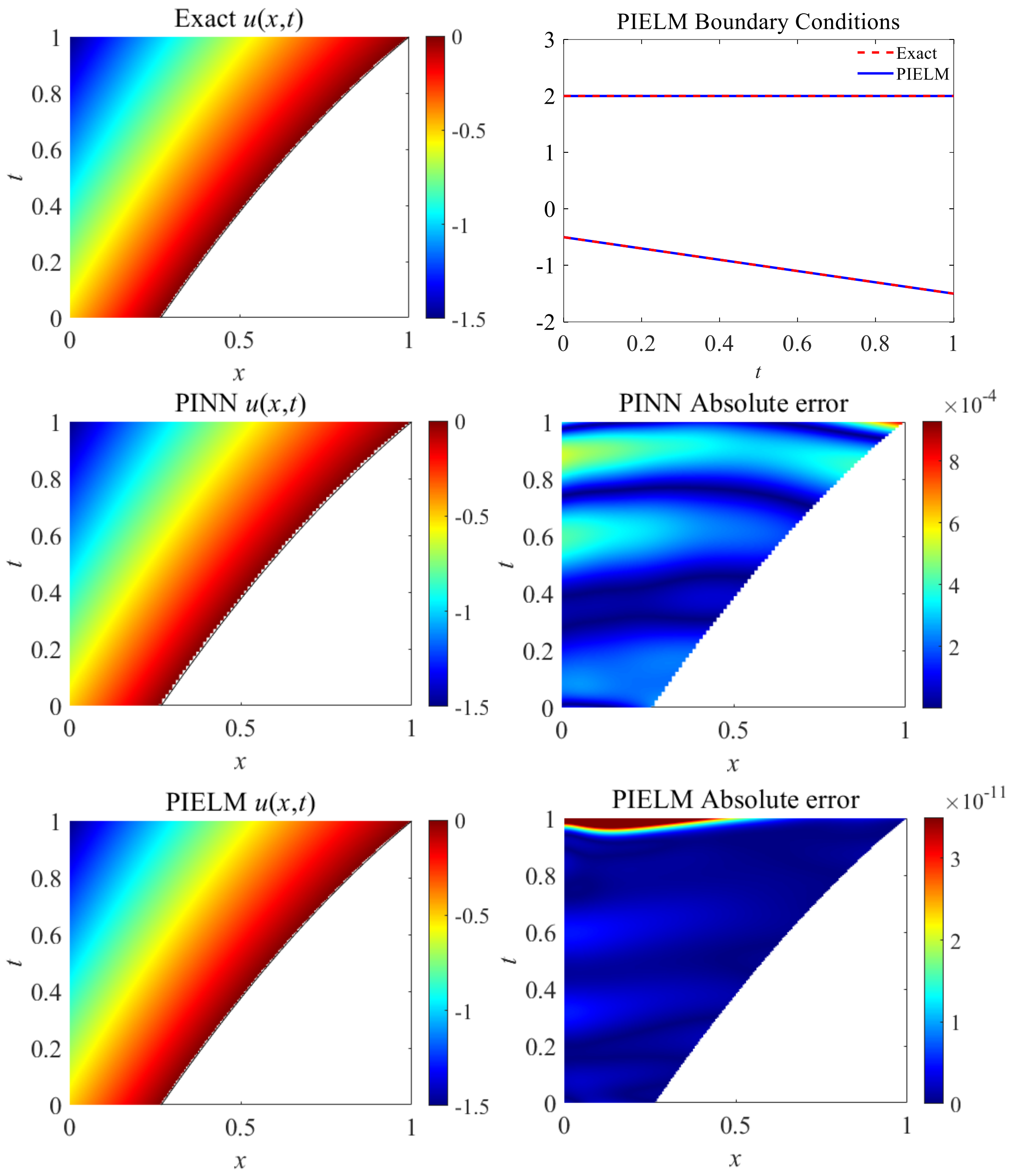

**Figure 5  Temperature distribution, boundary conditions, and absolute point error predicted by PINN and PIELM in Case 1**

As aforementioned, the input weights in the PIELM framework are generated randomly and are then fixed. The results predicted by PIELM may vary with the generated input weights, and the robustness of PIELM for inverse Stefan problem should be discussed. Table 3 summarizes the relative $L_2$ errors and training times of five repeated calculations by PIELM. Despite the random input weights, the results are rather stable and the robustness can then be confirmed.

**Table 3 Summary of relative L2 errors and training times**

| Test number | 1 | 2 | 3 | 4 | 5 |
|---|---|---|---|---|---|
| $L_2$ | 2.08e-11 | 6.85e-11 | 2.91e-11 | 4.93e-11 | 4.64e-11 |
| Training time (s) | 1.40 | 1.47 | 1.15 | 1.14 | 1.15 |

### *4.2. Case 2: One-dimensional Two-phase Inverse Stefan Problems*

The one-dimensional two-phase inverse Stefan problem in Wang and Perdikaris (2021) can be formulated as

$$\frac{\partial u_1}{\partial t} - 2\frac{\partial^2 u_1}{\partial x^2} = 0,\ \ 0 \le x \le s(t) \text{ and } 0 \le t \le 1 \tag{42}$$

$$\frac{\partial u_2}{\partial t} - \frac{\partial^2 u_2}{\partial x^2} = 0,\ s(t) \le x \le 2 \text{ and } 0 \le t \le 1 \tag{43}$$

$$u_1(x,0) = 2e^{\frac{1-2x}{4}} - 2,\quad 0 \le x \le s(0) \tag{44}$$

$$u_2(x,0) = e^{\frac{1-2x}{2}} - 1,\quad \mathrm{s}(0) \le x \le 2 \tag{45}$$

$$u_1(x,1) = 2e^{\frac{3-2x}{4}} - 2,\ \ 0 \le x \le s(1) \tag{46}$$

$$u_2(x,1) = e^{1.5-x} - 1,\ \ s(1) \le x \le 2 \tag{47}$$

$$u_1(s,t) = 0,\ \ 0 \le t \le 1 \tag{48}$$

$$u_2(s,t) = 0,\ \ 0 \le t \le 1 \tag{49}$$

$$-2\frac{\partial u_1}{\partial x}(s,t) + \frac{\partial u_2}{\partial x}(s,t) = 1,\ \ 0 \le t \le 1 \tag{50}$$

$$s(t) = t + \frac{1}{2} \tag{51}$$

The solution for this Stefan problem is

$$u_1(x,t) = 2e^{\frac{2t-2x+1}{4}} - 2, 0 \le x \le s(t) \tag{52}$$

$$u_2(x,t) = e^{\frac{2t-2x+1}{2}} - 1,\ s(t) \le x \le 2 \tag{53}$$

$$u_1(0,t) = 2e^{\frac{2t+1}{4}} - 2,\ 0 \le t \le 1 \quad (54)$$

$$u_2(2,t) = e^{\frac{2t-3}{2}} - 1,\ 0 \le t \le 1 \quad (55)$$

The predicted solutions by PINN and PIELM for this inverse Stefan problem are depicted in Figure 6. It can be found that the point-wise error of temperature distribution for PINN is in the order of $10^{-3}$, while the error for PIELM is less than $4\times10^{-6}$. Compared to PINN, the solution accuracy is improved more than 1000 times using PIELM in terms of the relative $L_2$ error, and training efficiency is improved more than 400 times, as shown in Table 2.

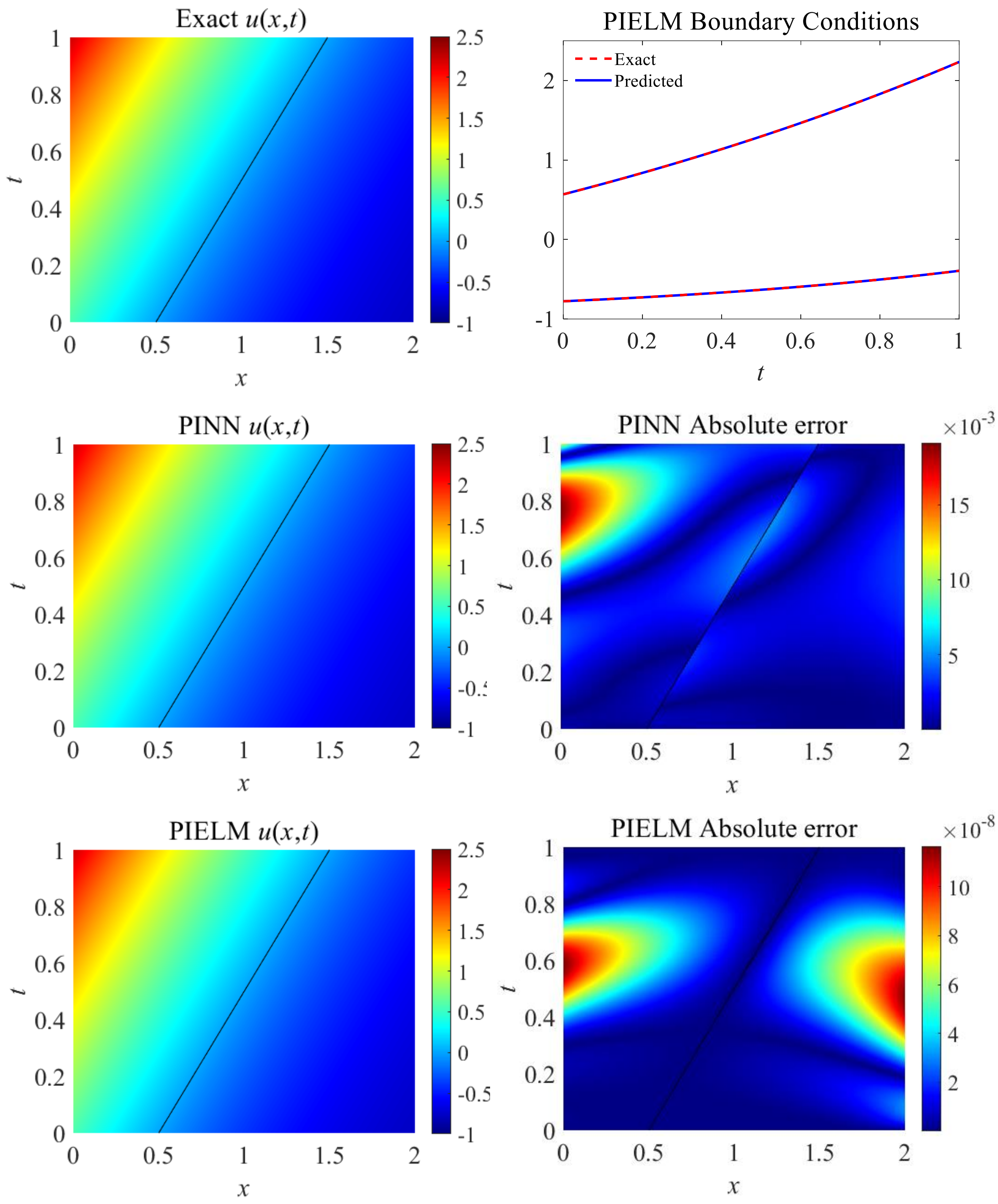

**Figure 6 Temperature distribution, boundary conditions, and absolute point error predicted by PINN and PIELM in Case 2**

### *4.3. Case 3: Two-dimensional One-phase Inverse Stefan Problems*

After showing the one-dimensional cases, a two-dimensional one-phase inverse Stefan problem is then analyzed by the PIELM method. As reported in Wang and Perdikaris (2021), the governing PDE as well as initial and boundary conditions can be expressed as

$$\frac{\partial u}{\partial t}-\frac{\partial^2 u}{\partial x^2}-\frac{\partial^2 u}{\partial y^2}=0,\ \ 0\le x\le s(y,t),\ \ 0\le y\le 1 \text{ and } 0\le t\le 1 \tag{56}$$

$$u(x,y,0)=e^{-x+0.5y+0.5}-1,\ \ 0\le x\le s(y,0) \text{ and } 0\le y\le 1 \tag{57}$$

$$u(x,y,1)=e^{-x+0.5y+1.75}-1,\ \ 0\le x\le s(y,1) \text{ and } 0\le y\le 1 \tag{58}$$

$$u(s,y,t)=0, 0\le y\le 1 \text{ and } 0\le t\le 1 \tag{59}$$

$$\frac{\partial u}{\partial x}(s,y,t)-\frac{1}{2}\frac{\partial u(s,y,t)}{\partial y}+\frac{5}{4}=0, 0\le y\le 1 \text{ and } 0\le t\le 1 \tag{60}$$

$$s(t)=\frac{5}{4}t+\frac{1}{2}y+\frac{1}{2} \tag{61}$$

The temperature distribution to be determined for this case is

$$u(x,y,t)=e^{\frac{5}{4}t-x+\frac{1}{2}y+\frac{1}{2}}-1,\ \ 0\le x\le s(t) \tag{62}$$

$$u(x,0,t)=e^{\frac{5}{4}t-x+\frac{1}{2}}-1 \tag{63}$$

$$u(x,1,t)=e^{\frac{5}{4}t-x+1}-1 \tag{64}$$

$$u(0,y,t)=e^{\frac{5}{4}t+\frac{1}{2}y+\frac{1}{2}}-1 \tag{65}$$

Figures 7 and 8 show the temperature distributions and absolute point errors predicted by PINN and PIELM at $t$ = 0.2s, 0.4s, 0.6s, and 0.8s, respectively. Figure 9 shows the predictions of PIELM for the boundary conditions. Both PINN and PIELM can provide relatively accurate solutions for the inverse Stefan problem in this case. However, in terms of point error, the accuracy of PIELM is $10^4$ times higher than that of PINN, reaching the $10^{-5}$ order of magnitude. In terms of training time, PIELM only requires 12 seconds for training. The comparison between $L_2$ error and training time further demonstrates the accuracy and efficiency of PIELM in the two-dimensional inverse Stefan problem.

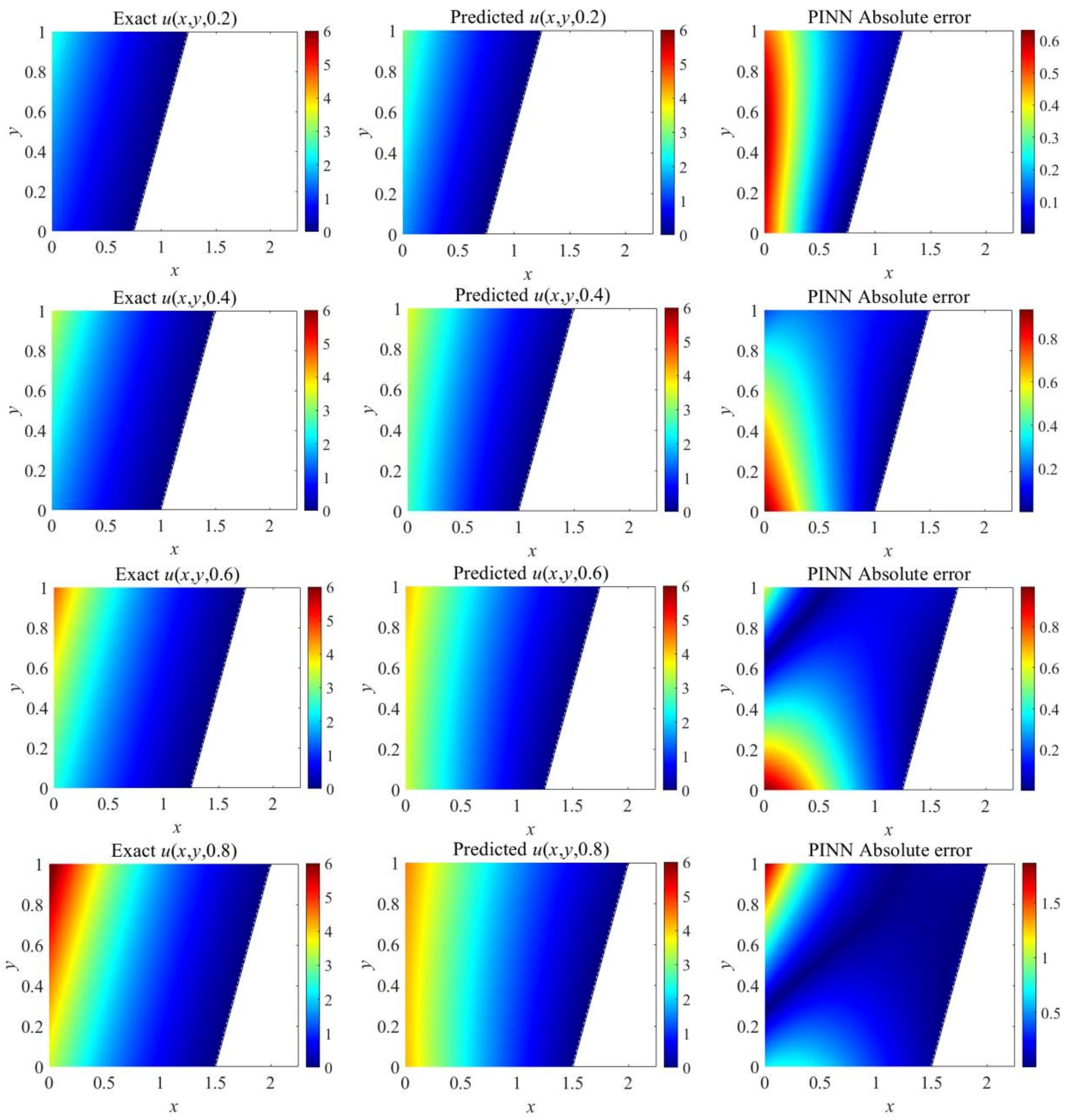

**Figure 7　Temperature distribution and absolute point-wise error predicted by PINN in Case 3**

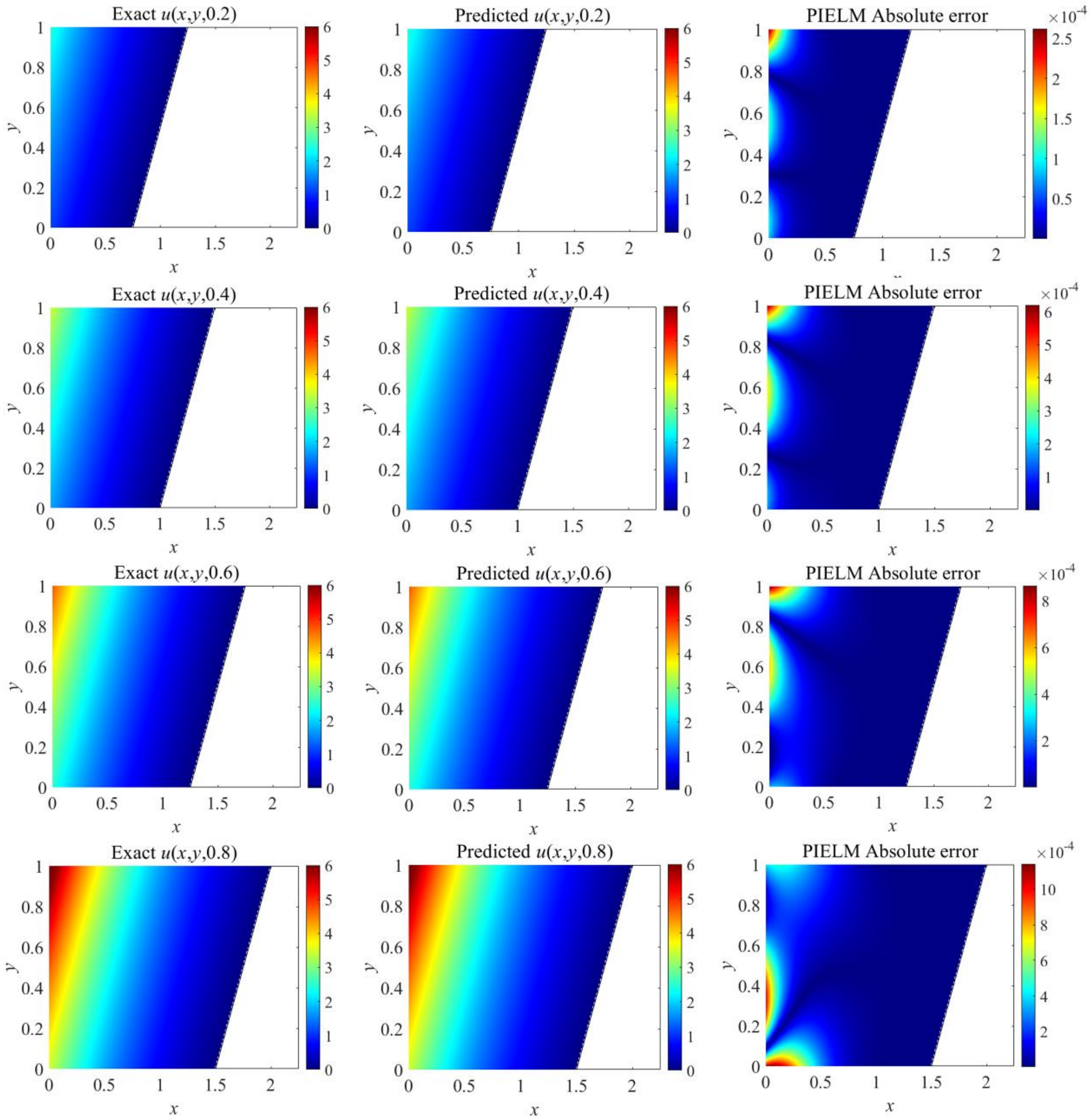

**Figure 8 Temperature distribution and absolute point-wise error predicted by PIELM in Case 3**

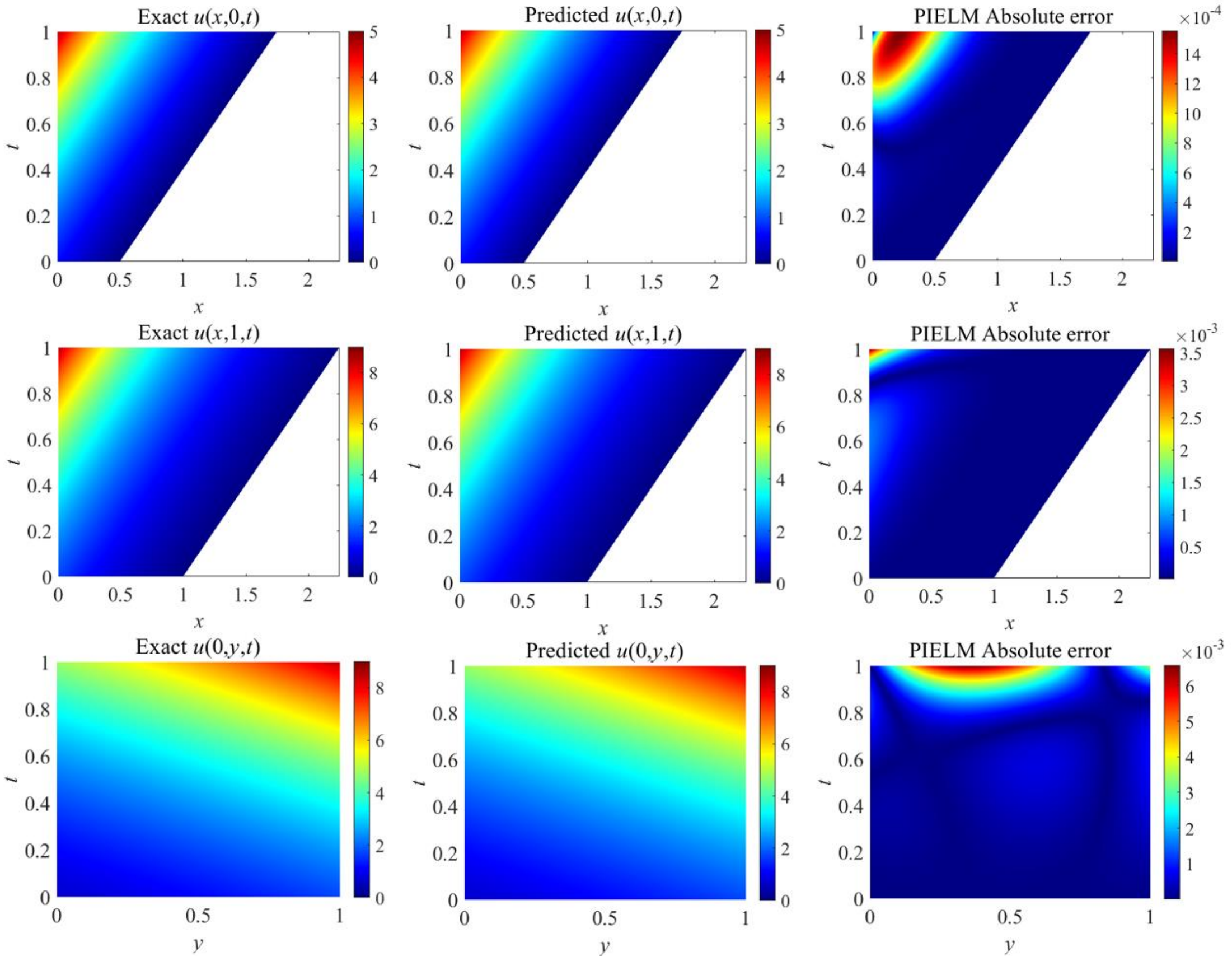

**Figure 9 Boundary conditions predicted by PIELM in Case 3**

### *4.4. Case 4: Two-dimensional Two-phase Inverse Stefan Problems*

In the final case, we want to test the performance of PIELM by a more challenging two-dimensional two-phase Stefan problem defined by Eqs. (66)~(79):

$$\frac{\partial u_1}{\partial t}-2\left(\frac{\partial^2 u_1}{\partial x^2}+\frac{\partial^2 u_1}{\partial y^2}\right)=0,\ 0\le x\le s(y,t),\ 0\le y\le 1 \text{ and } 0\le t\le 1 \tag{66}$$

$$\frac{\partial u_2}{\partial t}-\left(\frac{\partial^2 u_2}{\partial x^2}+\frac{\partial^2 u_2}{\partial y^2}\right)=0,\ s(y,t)\le x\le 2,\ 0\le y\le 1 \text{ and } 0\le t\le 1 \tag{67}$$

$$u_1(x,y,0)=2e^{\frac{1-2x}{4}}-2,\ 0\le x\le s(y,0) \text{ and } 0\le y\le 1 \tag{68}$$

$$u_2(x,y,0)=e^{\frac{1-2x}{2}}-1,\ s(y,0)\le x\le 2 \text{ and } 0\le y\le 1 \tag{69}$$

$$u_1(0,y,t)=2e^{\frac{2t+1}{4}}-2,\ 0\le y\le 1 \text{ and } 0\le t\le 1 \tag{70}$$

$$u_2(2,y,t)=e^{\frac{2t-3}{2}}-1,\ 0\le y\le 1 \text{ and } 0\le t\le 1 \tag{71}$$

$$u_1(x,0,t)=2e^{\frac{2t-2x+1}{4}}-2,\ 0\le x\le s(0,t)\text{ and }0\le t\le 1 \quad (72)$$

$$u_2(x,0,t)=e^{\frac{2t-2x+1}{2}}-1,\ s(0,t)\le x\le 2\text{ and }0\le t\le 1 \quad (73)$$

$$u_1(x,1,t)=2e^{\frac{2t-2x+1}{4}}-2,\ 0\le x\le s(1,t)\text{ and }0\le t\le 1 \quad (74)$$

$$u_2(x,1,t)=e^{\frac{2t-2x+1}{2}}-1,\ s(1,t)\le x\le 2\text{ and }0\le t\le 1 \quad (75)$$

$$s(y,0)=\frac{1}{2},\ 0\le y\le 1 \quad (76)$$

$$u_1(s,y,t)=0,\ 0\le y\le 1\text{ and }0\le t\le 1 \quad (77)$$

$$u_2(s,y,t)=0,\ 0\le y\le 1\text{ and }0\le t\le 1 \quad (78)$$

$$-2\left(\frac{\partial u_1(s,y,t)}{\partial x}-\frac{\partial u_1(s,y,t)}{\partial y}\frac{\partial s(y,t)}{\partial y}\right)+\left(\frac{\partial u_2(s,y,t)}{\partial x}-\frac{\partial u_2(s,y,t)}{\partial y}\frac{\partial s(y,t)}{\partial y}\right)-\frac{\partial s(y,t)}{\partial t}=0,\ 0\le y\le 1\text{ and }0\le t\le 1 \quad (79)$$

The exact solution can be expressed as

$$u_1(x,y,t)=2e^{\frac{2t-2x+1}{4}}-2,\ 0\le x\le s(y,t) \quad (80)$$

$$u_2(x,t)=e^{\frac{2t-2x+1}{2}}-1,\ s(y,t)\le x\le 2 \quad (81)$$

$$s(y,t)=t+\frac{1}{2} \quad (82)$$

Figures 10 and 11 show the temperature distribution and absolute point-wise error of PINN and PIELM for the two-dimensional two-phase inverse Stefan problem at $t$= 0.2s, 0.4s, 0.6s, and 0.8s, respectively. The $L_2$ error and training time are shown in Table 2. Compared with PINN, PIELM reduces the training error by $10^3$ and increases the training time by 20 times. This further demonstrates the high accuracy and speed of PIELM in solving inverse Stefan problems.

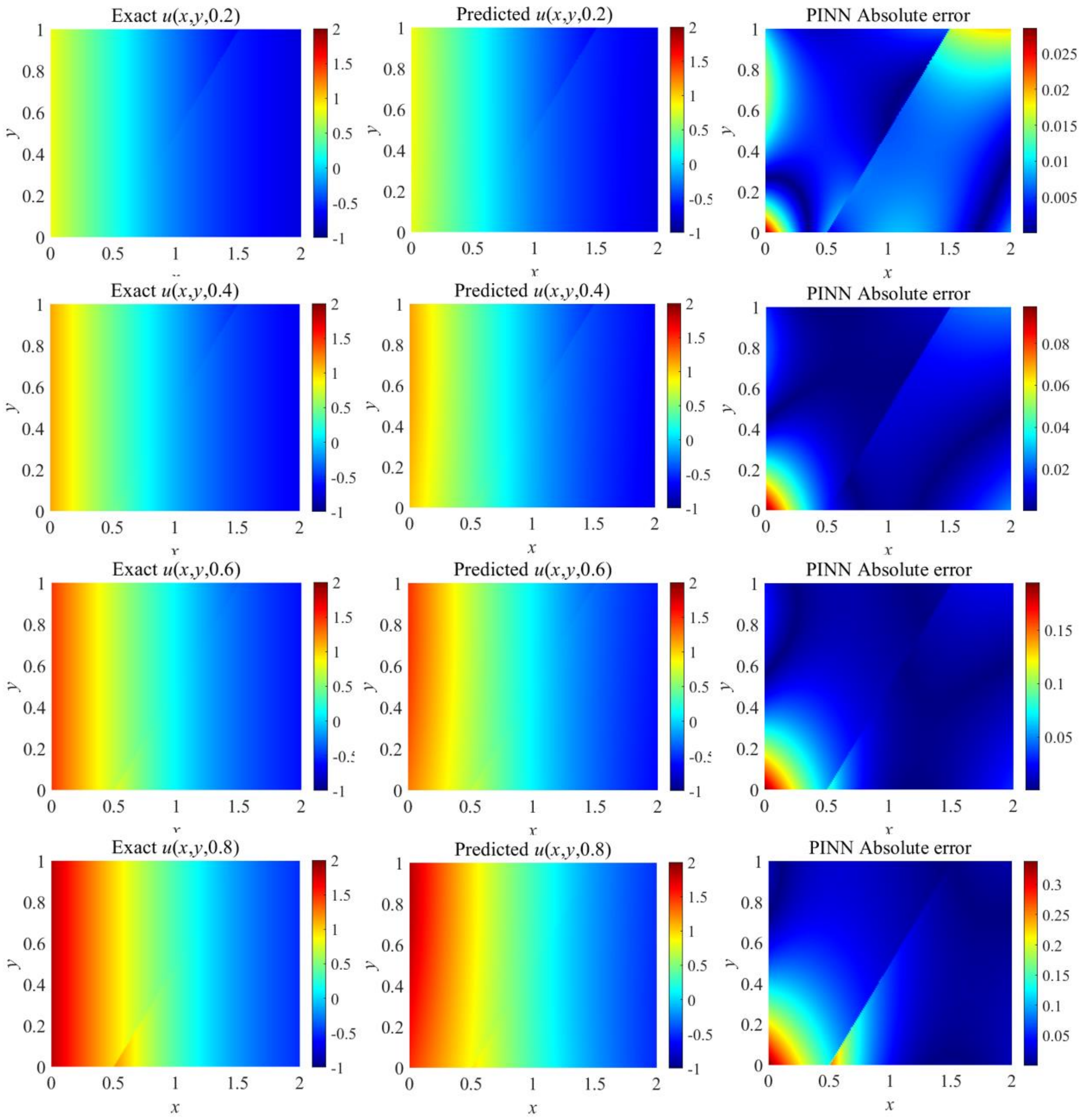

**Figure 7 Temperature distribution and absolute point-wise error predicted by PINN in Case 4**

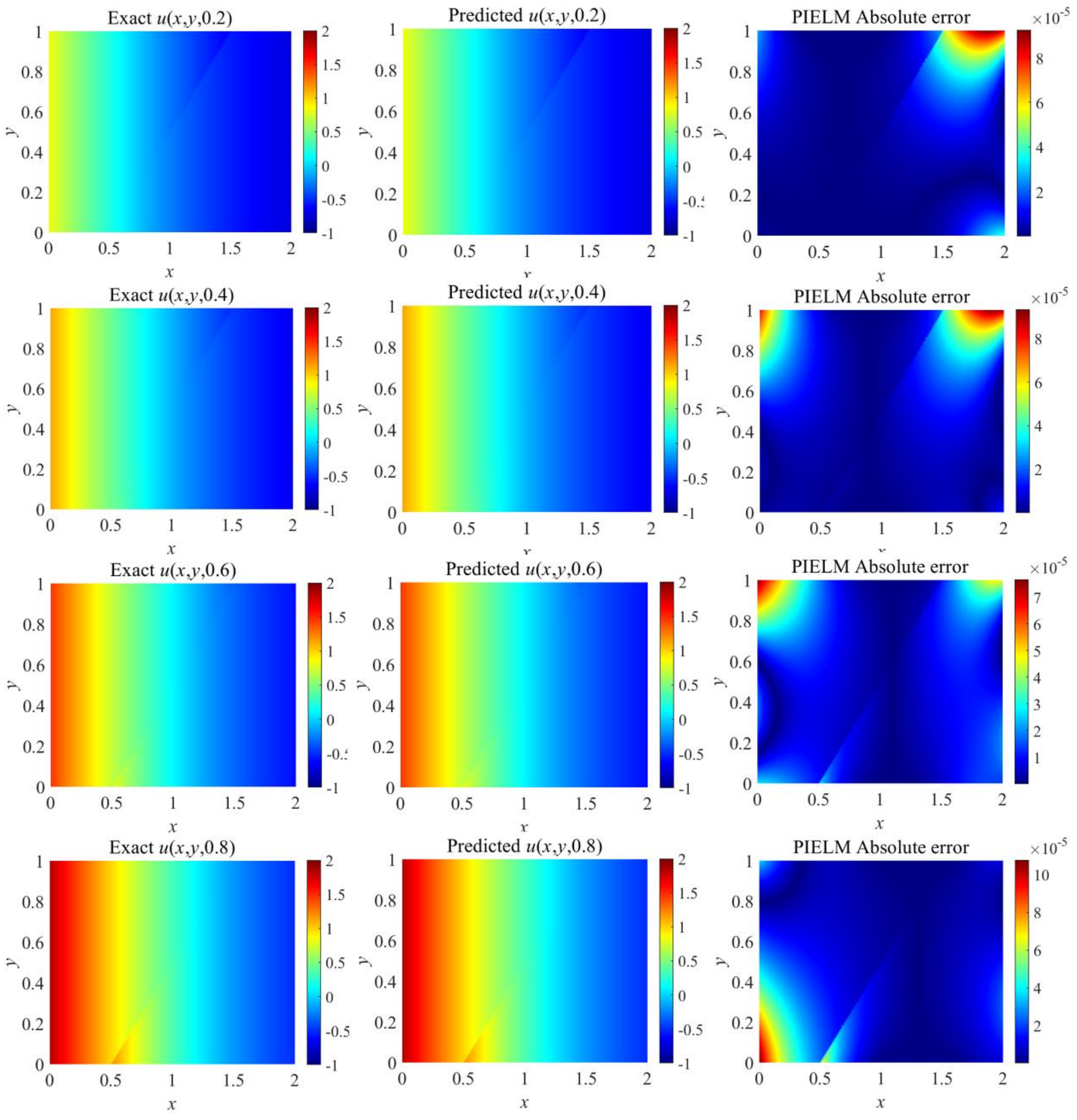

**Figure 7 Temperature distribution and absolute point-wise error predicted by PIELM in Case 4**

# 5. Conclusions

This paper proposes a physics-informed extreme learning machine (PIELM) framework and applies it to solving inverse Stefan problems. Compared to traditional PINN, PIELM employs a single-hidden-layer ELM network architecture, eliminating the need for deep neural networks and gradient descent training. This significantly enhances computational efficiency while achieving high improvement in prediction accuracy. Four numerical examples validate that PIELM improves 3~9 orders of magnitude over PINN regarding the relative $L_2$ error, while reducing 99.5% computation time for one-dimensional cases and 94% for the two-dimensional. These results demonstrate superior efficiency, precision, and robustness of PIELM, showing broad potential applications to engineering problems such as permafrost phase transition analysis, metal processing optimization, oil and gas extraction, and biological tissue modeling.

## Acknowledgement

We acknowledge the funding support from Natural Science Foundation of Shandong Province (ZR2024LZN002).

## Data Available Statement

Data are available from the corresponding author upon reasonable request.

## Conflict of Interests

The author declares that there is no known conflict of interests.